\documentclass[10pt,twocolumn,letterpaper]{article}

\usepackage[pagenumbers]{cvpr} 

\usepackage{graphicx}
\usepackage{amsmath}
\usepackage{amssymb}
\usepackage{booktabs}
\usepackage{epigraph} 
\usepackage{multicol}
\usepackage{multirow}
\usepackage{comment}
\usepackage{float}
\usepackage{tabu}
\usepackage{pifont}
\newcommand{\cmark}{\ding{51}}%
\newcommand{\xmark}{\ding{55}}%
\usepackage[dvipsnames]{xcolor}
\usepackage[pagebackref,breaklinks,colorlinks]{hyperref}

\usepackage[capitalize]{cleveref}
\crefname{section}{Sec.}{Secs.}
\Crefname{section}{Section}{Sections}
\Crefname{table}{Table}{Tables}
\crefname{table}{Tab.}{Tabs.}

\newcommand{\app}[0]{{Appendix}}

\newcommand\blfootnote[1]{%
  \begingroup
  \renewcommand\thefootnote{}\footnote{#1}%
  \addtocounter{footnote}{-1}%
  \endgroup
}

\makeatletter
\newcommand\notsotiny{\@setfontsize\notsotiny\@vipt\@viipt}
\makeatother

\def\cvprPaperID{4399} 
\def\confName{CVPR}
\def\confYear{2023}

\begin{document}

\title{AutoAD: Movie Description in Context}


\author{
Tengda Han$^{1*}$ \quad Max Bain$^{1*}$ \quad
Arsha Nagrani$^{1\dagger}$ \quad G\"ul Varol$^{1,2}$ \quad Weidi Xie$^{1,3}$ \quad Andrew Zisserman$^1$\\
{\small$^1$Visual Geometry Group, University of Oxford}\\ \quad
{\small$^2$LIGM, \'Ecole des Ponts, Univ Gustave Eiffel, CNRS}
\quad{\small$^3$CMIC, Shanghai Jiao Tong University}\\
{\small\url{https://www.robots.ox.ac.uk/vgg/research/autoad/}}
}

\maketitle

\begin{abstract}
The objective of this paper is an automatic Audio Description (AD) model that ingests movies and outputs AD in text form.
Generating high-quality movie AD is challenging due to the
dependency of the descriptions on context, 
and the limited amount of training data available.
In this work, we leverage the power of pretrained foundation models, such as GPT and CLIP, and only train a mapping network that bridges the two models for visually-conditioned text generation. 
In order to obtain high-quality AD, we make the following four contributions: 
(i)~we incorporate context from the movie clip, AD from previous clips, as well as the subtitles;
(ii)~we address the lack of training data by pretraining on large-scale datasets, where visual or contextual information is unavailable, e.g.\ text-only AD without movies or visual captioning datasets without context; 
 (iii)~we improve on the currently available AD datasets, 
by removing label noise in the MAD dataset, and adding character naming information; and
(iv) we obtain strong results on the movie AD task compared with previous methods.
\end{abstract}

\blfootnote{$*$: equal contribution.
$\dagger$: also at Google Research}

\vspace{-8mm}
\section{Introduction}
\label{sec:intro}
\renewcommand{\epigraphflush}{flushleft}
\renewcommand{\epigraphsize}{\small}
\setlength{\epigraphwidth}{0.43\textwidth}
{\scriptsize
\epigraph{\textit{That of all the arts, the most important for us is the cinema.}\\\hspace{155pt}\textit{Vladimir Lenin}}
{}
}
\vspace{-3mm}

\begin{figure}[t!]
    \centering
    \includegraphics[width=0.49\textwidth]{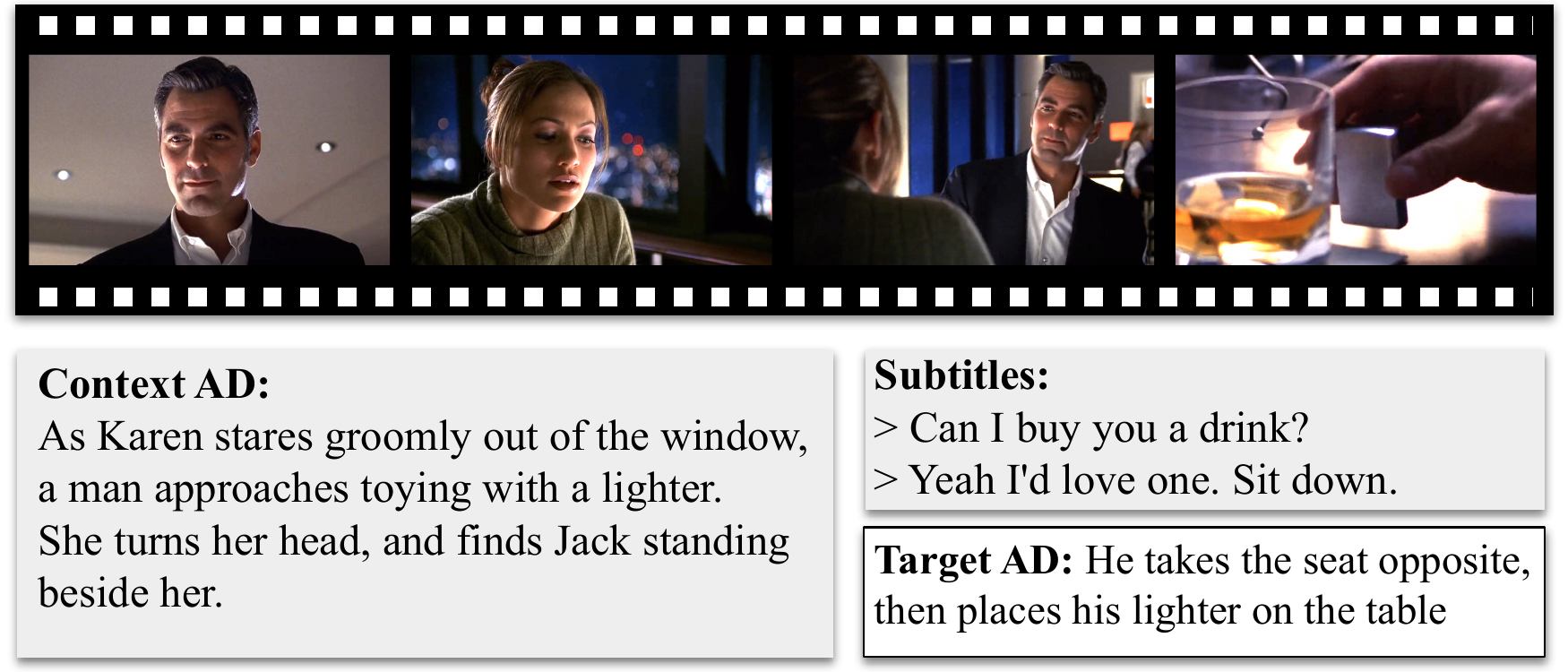}
    \vspace{-6mm}
    \caption{\textbf{Movie audio description (AD)} consists of sentences describing movies for the visually impaired. Note how it is heavily influenced by various types of context -- the visual frames, the previous AD, and the subtitles of the movie. 
    }
    \vspace{-5mm}
    \label{fig:teaser}
\end{figure}

One of the long-term aims of computer vision is to understand long-form feature films. There has been steady progress towards this aim with the identification of characters by their face and voice~\cite{bojanowski2013act,Brown21,Everingham06a,Tapaswi12,PersonSearch18}, the recognition of their actions and inter-actions~\cite{Laptev08,marszalek09,PatronPerez10,vondrick2016anticipating}, of their relationships~\cite{Kukleva_2020_CVPR}, and 3D pose~\cite{Angjoo22}. However, this is still a long way away from story understanding.
Movie {\em Audio Description (AD)}, 
the narration describing visual elements in movies, 
provides a means to evaluate current movie understanding capabilities.
AD was developed to aid visually impaired audiences, and is typically generated by experienced annotators. 
The amount of AD on the internet is growing due to more societal support for visually impaired communities and its inclusion is becoming an emerging legal requirement.

AD differs from image or video captioning in several significant respects~\cite{youdescribe}, bringing its own challenges.
First, AD provides dense descriptions of important visual elements {\em over time}.
Second, AD is always provided on a separate soundtrack to the original audio track and is highly {\em complementary} to it.
It is complementary in two ways: it does not need to provide descriptions of events that can be understood from the soundtrack alone (such as dialogue and ambient sounds), and it is constrained in time to intervals that do not overlap with the dialogue.
Third, unlike dense video captioning, AD aims at {\em storytelling};
therefore, it typically includes factors like a character's name, emotion, and action descriptions.

In this work, our objective is automatic AD generation -- a model that takes continuous movie frames as input and outputs AD in text form. 
Specifically, we generate text given a temporal interval of an AD, and evaluate its quality by comparing with the ground-truth AD.
This is a relatively unexplored task in the vision community with previous work targeting ActivityNet videos~\cite{wang2021toward}, a very different domain to long-term feature films with storylines, and the LSMDC challenge~\cite{rohrbach2015lsmdc}, where the descriptions and character names are treated separately.

As usual, one of the challenges holding back progress is the lack of suitable training data.
Paired image-text or video-text data that is available at scale, such as alt-text~\cite{clip2021,sharma2018cc} or stock footage with captions~\cite{Bain21}, 
does not generalize well to the movie domain~\cite{bain2022clip}.
However, collecting high-quality data for movie understanding is also difficult.
Researchers have tried to hire human annotators to describe video clips~\cite{chen2015cococap,xu2016msrvtt,krishna2017anetcaption} but this does not scale well.
Movie scripts, books and plots have also been used as learning signals~\cite{bojanowski2013act,sigurdsson2016charades,Zhu2015AligningBA} but they do not ground on vision closely and are limited in number.

In this paper we address the AD and training data challenges by -- Spoiler Alert -- developing a model that uses temporal context
together with a visually conditioned generative language model, while providing new and cleaner sources of training data.
To achieve this, we leverage the strength of large-scale language models (LLMs), like GPT~\cite{gpt2019}, and vision-language models, like CLIP~\cite{clip2021},
and integrate them into a video captioning pipeline that can be effectively trained with AD data.

Our contributions are the following: 
(i) inspired by ClipCap~\cite{mokady2021clipcap} we propose a model that is effectively able to leverage both temporal context (from previously generated AD) and dialogue context (in particular the names of characters) to improve AD generation. This is done by bridging foundation models with lightweight adapters to integrate both types of context; 
(ii) we address the lack of large-scale training data for AD by pretraining components of our model on partially missing data which are typically available in large quantities e.g.\ text-only AD without movie frames, or visual captioning datasets without multiple sentences as context; 
(iii) we propose an automatic pipeline for collecting AD narrations at scale using speaker-based separation; and finally 
(iv) we show promising results on automatic AD, as seen from both qualitative and quantitative evaluations, 
and also achieve impressive \emph{zero-shot} results on the LSMDC multi-description benchmark comparable to the finetuned state-of-the-art.

\section{Related Works}
\label{sec:related_work}


\noindent \textbf{Image Captioning.}
Image captioning is a long-standing problem in computer vision~\cite{Chen2014LearningAR,donahue2015long,karpathy2015deep,kiros2014unifying,Lu2018NeuralBT,anderson2018bottomup,chen2015cococap}.
Early pioneering works learn to associate images and words within a limited vocabulary and a set of images~\cite{barnard2001semantics,barnard2003matching,lavrenko2003model}.
Large-scale image captioning datasets have been collected by scraping images from the internet and their corresponding alt-texts with quality filters as a post-processing~\cite{sharma2018cc}. In doing so, strong joint image-text representations can be learned~\cite{clip2021}, and image captioning from raw pixels, with impressive results~\cite{yu2022coca,li2022blip}.
Recent work~\cite{mokady2021clipcap,nukrai2022text} learns a bridge between strong joint image-text representations (CLIP) and the natural language representation (GPT-2) for image captioning, obtaining promising results that generalise well across domains. In this work, we extend this approach to perform automatic AD
from videos.

\vspace{2pt}
\noindent \textbf{Video Captioning.}
Video captioning presents additional challenges due to the lack of quality large-scale video-text data and increased complexity from the temporal axis. Early video caption datasets~\cite{chen2011msvd,xu2016msrvtt} adopt manual annotations, a far from scalable collection method. ASR (automated speech recognition) from YouTube instructional videos is collected at scale for video-language datasets~\cite{miech2019howto100m}, but contains high levels of noise due to the weak correspondence between the narration and visual content. VideoCC~\cite{nagrani2022learning} transfers captions from images to videos, but this method is still limited by the existing seed image captioning dataset used.
Earlier video captioning models lack generalisation capabilities 
due to limited training data~\cite{Venugopalan2015SequenceTS,Park2019AdversarialIF}. Some recent methods~\cite{seo2022end,huang2020multimodal,luo2020univl} train on ASR from the HowTo100M dataset, while others expand image-text representations~\cite{tang2021clip4caption} to multiple frames.

A task more related to AD is that of dense video captioning~\cite{krishna2017dense}, which involves producing a number of captions and their corresponding grounded timestamps in the video. To enrich inter-task interactions, recent works for this task~\cite{chadha2020iperceive, chen2021towards, deng2021sketch, li2018jointly, mun2019streamlined, rahman2019watch, shen2017weakly, shi2019dense, wang2018bidirectional, wang2021end, zhou2018end} jointly train both a captioning and localization module. Our task differs in that the captions are: made with the intent to aid storytelling; specific to the movie domain; and complementary to the audio track.

\noindent \textbf{Visual Storytelling.} Most similar in vein to the AD task is visual storytelling~\cite{huang2016visual,Li2020VideoST,ravi2021aesop}, in which the goal is to generate coherent sentences for a sequence of video clips or images. LSMDC~\cite{rohrbach2017movie} proposes the multi-description task of generating captions for a set of clips from a movie, with character names anonymized. In contrast, movie AD takes as input a continuous long video and describes the visual happenings complementary to the story, characters, dialogue and audio. Most similar to our model is TPAM~\cite{yu2021transitional} which prompts a frozen GPT-2 with local visual features. Ours differs in that: (i) it is not restricted to local visual context but rather global by recurrently conditioning on previous outputs; and (ii) we additionally pretrain GPT on in-domain text-only AD data.

\noindent \textbf{Movie Understanding.} Previous works investigate storyline understanding by aligning movies to additional data sources such as plots~\cite{xiong2019graph,sun2022synopses}, books ~\cite{tapaswi2015book2movie,Zhu2015AligningBA},
scripts~\cite{papalampidi2019movie}, and YouTube summaries~\cite{bain2020cmd}. However, these sources are limited in number and often do not closely relate to the visual elements in the frame.
Using existing movie AD as the data source for videos is an emergent direction for movie understanding. LSMDC~\cite{rohrbach2015lsmdc}, M-VAD dataset~\cite{torabi2015mvad} and MPII-MD~\cite{rohrbach2015dataset}, gather AD and scripts from movies to provide captions for short video clips, several seconds in duration. QuerYD~\cite{oncescu2021queryd} provides high-quality textual descriptions for longer videos by scraping AD from YouDescribe~\cite{youdescribe}, an online community of AD contributors. Recently, the MAD dataset~\cite{soldan2022mad} collects movie AD at scale to provide dense textual annotations for movies with a focus on visual grounding task.

\vspace{2pt}
\noindent \textbf{Prompt Tuning and Adapters.}
Originally for language modelling,
prompt tuning is a lightweight approach to adapt a pretrained model
to perform a downstream task.
Early works~\cite{Brown2020,Lester2021,LiLiang2021,ju2022prompting}
learn prompt vectors that are shared within the targeted dataset and task.
A similar line of works to ours is 
\emph{visual-conditioned} prompt tuning,
in which the prompt vectors are conditioned on the visual inputs.
Visual-conditioned prompts are used for adapting
pretrained image-language models~\cite{Bahng2022,Jia2022},
and for few-shot learning~\cite{tsimpoukelli2021frozen,alayrac2022flamingo}.
Training lightweight feature adapters between pretrained vision and text encoders
is another approach to adapt pretrained models~\cite{gao2021clip, zhang2021tip}.
The adapter layers can also be inserted into the pretrained language model in 
an interleaved way~\cite{yang2022zero}. Our work adopts prompt tuning in order to condition a language generation model on visual information (frames), and textual context (subtitles and previous AD).

\begin{figure*}[t]
    \centering
    \includegraphics[width=0.99\textwidth]{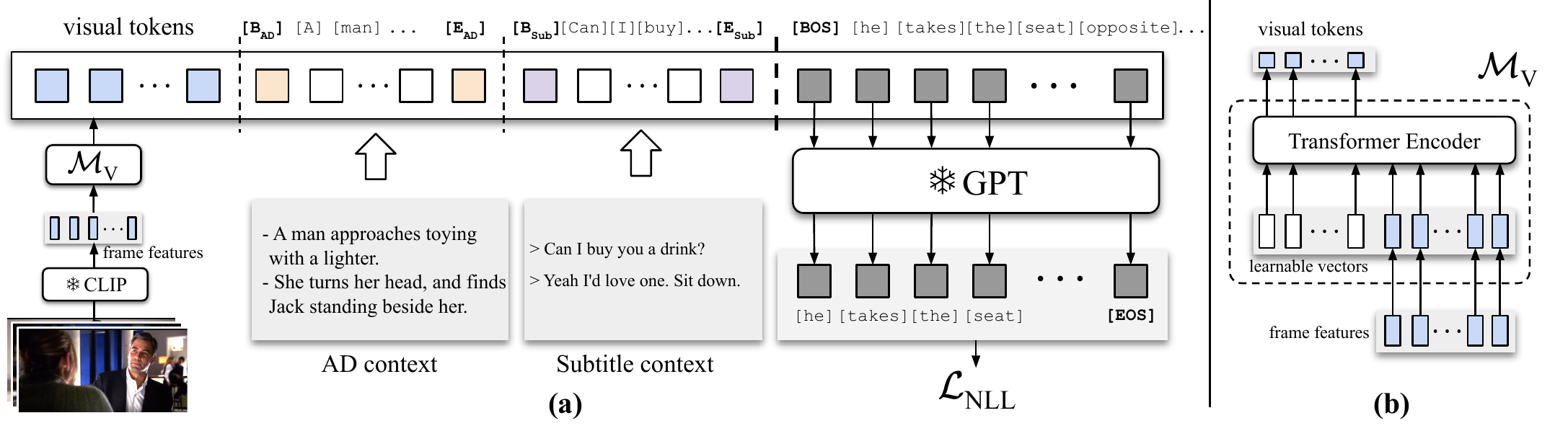}
    \vspace{-10pt}
    \caption{
    \textbf{(a) Overview of AutoAD:}
    AutoAD consists of a \textit{frozen} visual encoder (CLIP) and a \textit{frozen} LLM (GPT) for generating captions. We introduce a lightweight mapping network to map CLIP features into visual tokens, which are then combined with previous AD context and subtitle context, before being fed into the GPT model. 
    $\mathcal{M}_\text{V}$ refers to the visual mapping network,
    $[\tt{B_*}]$
    and $[\tt{E_*}]$ 
    denote the learnable special tokens for contextual AD and subtitle sequences.
    \textbf{(b) Detail of the {visual mapping network}:}
    A transformer encoder takes as input multiple frame features and outputs 
    a few visual tokens which are further fed to a text generation model.
    }
    \vspace{-5mm}
    \label{fig:arch_context}
\end{figure*}

\section{Method}
\label{sec:method}

Given a long-form movie $\mathcal{V}$ segmented into multiple short clips $\{ \mathbf{x}_1, \mathbf{x}_2, ..., \mathbf{x}_T \}$, our goal is to generate the audio description~(AD) in text form for every movie clip. 
Note that each movie clip is cut from the raw movie based on the timestamp $[t_{\text{start}}, t_{\text{end}}]$ given by the AD annotation.
Specifically, for the $i$-th movie clip consisting of multiple frames 
$\mathbf{x}_i = \{ \mathcal{I}_1, \mathcal{I}_2, ..., \mathcal{I}_N \}$, we aim to produce text $\mathcal{T}_i$ that describes the visual elements in such a way that helps the visually impaired follow the storyline.
To this purpose, an ideal AD generation system must be able to exploit the full contextual information leading up to the $i$-th movie clip. One method for this, which we adopt, is to use previous AD $\mathcal{T}_{t<i}$ and subtitles $\mathcal{S}_{t\leq i}$ to generate the text $\mathcal{T}_i$.
In the following sections,
we first give an overview of our visual captioning pipeline with prompt tuning (Sec.~\ref{sec:method:base}),
followed by our contextual components (Sec.~\ref{sec:method:temporal}),
and finally the pretraining methods with partial data (Sec.~\ref{sec:method:partialpt}).

\subsection{Visual Captioning with Prompt Tuning}
\label{sec:method:base}

In order to describe our method, we first present the typical pipeline for an image captioning model, and then detail how we extend this to ingest multiple frames and additional text context.
Given an image-caption pair $\{\mathcal{I}_i, \mathcal{C}_i\}$, where the caption consists of a sequence of language tokens $\mathcal{C}_i=\{c_1,c_2,...,c_k\}$,
the standard objective of an image captioning model is
to generate text tokens $\hat{\mathcal{C}}_i$ that are close to the target $\mathcal{C}_i$.
Technically, the captioning models are trained to maximize
the joint probability of predicting the ground-truth language tokens,
or equivalently minimize the following negative log-likelihood (NLL) loss,
\vspace{-2mm}
\begin{align}\vspace{-2mm}
    \mathcal{L}_{\text{NLL}}
    &= - {\log{p_{\theta}{(\mathcal{C}_i | \mathbf{h}_{\mathcal{I}_i})}}}
    = - {\log{p_{\theta}{(c_1,c_2,...,c_k | \mathbf{h}_{\mathcal{I}_i})}}} \nonumber 
    \label{eq:nll}
\vspace{-3mm}
\end{align}
where $\theta$ denotes the parameters of the model,
and $\mathbf{h}_{\mathcal{I}_i}$ denotes the extracted image features of $\mathcal{I}_i$.
Previous works like ClipCap~\cite{mokady2021clipcap} fit a powerful text generation model 
and visual encoding model into this image captioning pipeline.
Specifically, strong visual encoding models, such as CLIP~\cite{clip2021}, are used to
extract the visual features from the input image
$\mathbf{z}_i = f_{\text{CLIP}}(\mathcal{I}_i)$,
then a visual mapping network $\mathcal{M}_{\text{V}}$ is trained to map the visual features to
`prompt vectors' that adapt to the text generation model,
$\mathbf{h}_{\mathcal{I}_i} = \mathcal{M}_{\text{V}}(\mathbf{z}_i)$.
Finally these prompt vectors $\mathbf{h}_{\mathcal{I}_i}$ are fed to a pretrained text generation model, such as GPT~\cite{gpt2019}, for the captioning task.
We adapt this visual captioning pipeline, which uses pretrained feature extractor CLIP and langauge model GPT, for movie AD generation and propose key components that support contextual understanding.

\subsection{Benefiting from Temporal Context}
\label{sec:method:temporal}

Here, we describe how we extend this single-frame captioning model to include different forms of context, including multiple frames, previous AD text, and subtitles. 
Compared to image captioning where the annotation describes `what is in the image', movie AD describes the visual happenings in the scene that are relevant to the broader story -- often centered around events, characters and the interactions between them.
Factors like these cannot be accurately described from a static image alone and therefore a successful automatic AD system must utilize the context of prior events and character interactions.

To tackle these temporal dependencies,
we propose to include three components to incorporate the essential contextual information from movies:
(i) immediate visual context in the current movie clip (multiple frames),
(ii) the previous movie AD, and 
(iii) the movie subtitles.
The architecture of our model is shown in Fig.~\ref{fig:arch_context}.

\vspace{3pt}
\noindent \textbf{Multiple frames (immediate visual context).}
In contrast to the image captioning method,
the visual mapping network $\mathcal{M}_{\text{V}}$ takes as input multiple frame features from the
current movie clip $\mathbf{x}_i$ rather than a single image feature, 
and outputs prompt vectors for the movie clip,
\vspace{-3mm}
\begin{equation*}\vspace{-2mm}
    \mathbf{h}_{\mathbf{x}_i} = \mathcal{M}_{\text{V}}(\{ \mathbf{z}_1, ..., \mathbf{z}_N \});
\quad 
\mathbf{z}_i = f_{\text{CLIP}}(\mathcal{I}_i).
\vspace{-3pt}
\end{equation*}
In detail, the mapping network consists of a multi-layer transformer encoder that enables modelling 
temporal relations among multiple frame features, as shown in Fig~\ref{fig:arch_context}.

\vspace{3pt}
\noindent \textbf{Previous AD text.}
The sequence of events leading up to the present contain contextual information which are crucial for generating AD of current scene that helps the viewer follow the story. 
We input this contextual knowledge to our model in the form of the past ADs.
Specifically, our model takes the past $K$ movie ADs $\{ \mathcal{T}_{i-K}, ..., \mathcal{T}_{i-1} \}$ to generate the AD for the current clip.
The past movie ADs are a few sentences, which are first concatenated into a single paragraph,
then tokenized and converted to a sequence of word embeddings.
Inspired by the design of special tokens in language models,
we wrap the context AD embeddings with \emph{learnable} special tokens to indicate 
the beginning and end of the AD sequence. 
Formally, the contextual AD embedding is a sequence,
\vspace{-3mm}
\begin{equation}\vspace{-2mm}
    \mathbf{h}_{\text{AD}} = [
\texttt{B}_{\text{AD}};
\mathbf{h}_{\mathcal{T}_{i-K}}; ...; \mathbf{h}_{\mathcal{T}_{i-1}};
\texttt{E}_{\text{AD}}]
\label{eq:context_ad}
\end{equation}

\noindent where $\texttt{B}_{\text{AD}}$ and $\texttt{E}_{\text{AD}}$ 
are the learnable special tokens indicating the beginning and end, 
the symbol `;' denotes concatenation,
and $\mathbf{h}_{\mathcal{T}_{j}} \in \mathbb{R}^{n \times C}$ denotes the word embedding of the $j$-th movie ADs.

\vspace{3pt}
\noindent \textbf{Previous subtitles.} Our model also takes the movie subtitles as additional contextual information, which can be sourced either from the official movie metadata or automatically transcribed with an ASR model.
The character dialogues, contained with the subtitles, provide complementary information to movie description, including the character names, relationships and emotions.
Similar to the context ADs,
we concatenate multiple subtitle sentences into a single paragraph
and wrap them with learnable special tokens.
Practically, since the timing of movie AD does not overlap with the subtitles,
we take the most recent $L$ subtitles within a certain time range as the context,
\vspace{-2mm}
\begin{equation*}\vspace{-2mm}
\mathbf{h}_{\text{Sub}} = [
\texttt{B}_{\text{Sub}};
\mathbf{h}_{\mathcal{S}_{i-L}}; ...; \mathbf{h}_{\mathcal{S}_{i-1}};
\texttt{E}_{\text{Sub}}]
\end{equation*}
Due to the weak correlation between the subtitles and the visual elements in the scene,
we also experiment with a variant that only encodes the character names occurring in the recent subtitles.

\vspace{3pt}
\noindent \textbf{Summary.}
Overall, the movie AD for the current movie clip $\mathcal{T}_{\mathbf{x}_i}$ is generated 
by conditioning on all the previously described visual and contextual information using a pretrained GPT.
The conditional information is fed to GPT as prompt vectors as shown in Fig.~\ref{fig:arch_context}.
The model is trained with NLL loss,
\vspace{-1mm}
\begin{align}\vspace{-3mm}
    \mathcal{L}_{\text{NLL}}
    &= - {\log{p_{\Theta}{(
        \mathcal{T}_{\mathbf{x}_i} | \mathbf{h}_{\mathbf{x}_i},\mathbf{h}_{\text{AD}},\mathbf{h}_{\text{Sub}} 
        )}}}.
    \label{eq:nll-ours}
\end{align}

\noindent During training, we input the ground-truth past AD.
During inference, we experiment with two methods to incorporate the past AD:
an \textbf{oracle} setting where the \emph{ground-truth} past ADs are used in Eq.~\ref{eq:nll-ours} to generate the current AD, and a
\textbf{recurrent} setting where the \emph{predicted} past ADs are used instead.

\begin{figure*}
    \centering
    \hfill
    \includegraphics[width=0.95\textwidth]{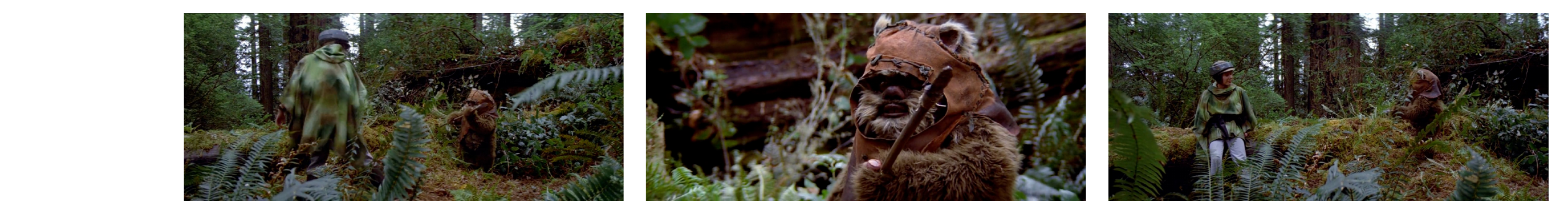}
\\
\tiny

\setlength{\tabcolsep}{2pt}
\resizebox{2.1\columnwidth}{!}{%
\begin{tabu}{@{}l|l|l|l@{}}
\hline 
\rowfont{\color{gray}} \textbf{Manual Verification*}           &  She stands and the little warrior takes in her size, about twice his own. &  As she steps past him, he defensively grips his spear &  Leia sits on a moss covered log. \\ \hline
\textbf{MAD-v1~\cite{soldan2022mad}} & {\color{red}Angola,} she stands {\color{red}in} the Little Warrior, takes in her size about twice his own.   & As she steps past {\color{red}me.} Defensively grips his spear.    & {\color{red}I'm not gon na.} {\color{red}Leah} sits on a Moss covered log. \\
\textbf{MAD-v2 (ours)}       & She stands and the little warrior takes in her size about twice his own.  & As she steps past him, he defensively grips his spear. &  Leia sits on a moss-covered log. \\ 
\hline 
\end{tabu}
}
\vspace{-3mm}
\caption{\textbf{Qualitative comparison of MAD annotations.} We compare the original MAD-v1~\cite{soldan2022mad} and our proposed MAD-v2. Note MAD-v1's erroneous transcriptions of AD and dialogue leakage (highlighted in {\color{red}red text}). The samples are taken from Star Wars VI: Return of the Jedi (1983)~\cite{starwars83}. *We verify this example by manually transcribing the AD narration from the audio track.
}
\vspace{-4mm}
\label{fig:mad-anno}
\end{figure*}

\subsection{Pretraining with Partial Data}
\label{sec:method:partialpt}

A major challenge for generating AD is the lack of training data,
since the model requires the corresponding visual, textual and contextual data to all be jointly trained.
However since our model is modular, components of it can be pretrained with \emph{partial data}
-- when a certain type of data is missing, the remaining modules can still be trained.
We experiment with partial-data pretraining under two settings:
visual-only pretraining and AD-only pretraining.

\vspace{3pt}
\noindent \textbf{Visual-only Pretraining.}
In the absence of contextual data,
the visual mapping network can be pretrained with abundant
image captioning or (short) video captioning datasets.
In this case, the context modules (both contextual AD and subtitles) are deactivated.
The training objective of Eq.~\ref{eq:nll-ours} is turned into 
$\mathcal{L} = - {\log{p_{\Theta}{(
    \mathcal{T}_{\mathbf{x}_i} | \mathbf{h}_{\mathbf{x}_i}
    )}}}$
for visual-only pretraining.
Note that the language model is kept frozen here since 
we find image/video captioning datasets have a clear domain gap
with movie AD in both the vision and text modalities.

\vspace{3pt}
\noindent \textbf{AD-only Pretraining.}
\label{sec:ad-pretrain}
Movie AD datasets with corresponding visual information (\eg~frames or frame features)
are limited at scale due to potential copyright issues.
However, abundant \emph{text-only} movie ADs are available online as described in Sec.~\ref{sec:audiovault_textonly}.
In the absence of visual data,
the contextual AD module and the language model can still be pretrained.
The training objective in this case becomes 
$\mathcal{L} = - {\log{p_{\Theta}{(
    \mathcal{T}_{\mathbf{x}_i} | \mathbf{h}_{\text{AD}}
    )}}}$,
which is similar to training a story completion objective~\cite{mostafazadeh2016corpus}
by finetuning GPT on \emph{text-only} movie AD data but with a few additional special tokens.
This text-only movie AD pretraining is also related to~\cite{gururangan2020don},
which shows a second stage of language model pretraining on in-domain data improves downstream performance.

\section{Denoising MAD Dataset}
\vspace{-1mm}
\label{sec:mad}

Our main objective is to generate movie audio descriptions.
For this goal, the model is trained on the MAD training set~\cite{soldan2022mad},
a dataset of AD caption-video clip pairs from 488 movies. MAD provides the video data in the form of CLIP visual features in order to avoid copyright restrictions.
The AD annotations for each movie are automatically collected from AudioVault\footnote{\url{https://audiovault.net}\label{fn:audiovault}}, 
a large open-source database of audio files containing the full-length original movie track mixed with the AD narrator's voice. The MAD authors transcribe a subset of this data using ASR, and also have access to the official DVD subtitles. 
Their automated method then uses \textit{text-based} speaker separation of the transcribed audio by using subtitles to know when dialogue is present, and assuming all other speech is AD.

This however introduces \emph{significant noise} because 
(i) the outdated ASR model results in erroneous transcriptions; and (ii) official DVD subtitles are not exhaustive of all speech in the movie and thus such a method frequently misidentifies character dialogue as AD narration
(an example is provided in Fig.~\ref{fig:mad-anno}). Further, obtaining official subtitles from DVDs presents additional challenges when collecting this data at scale.

We propose an improved automated data collection method for AD, requiring only the audio track as input (no DVD subtitles), that tackles both issues by using \textit{audio-based} speaker separation and an improved ASR model.
We then use this method to collect improved annotations for the MAD dataset.
Briefly, taking the mixed audio containing both AD narrations and original movie sound track as input,
our automated AD collection pipeline contains five stages:
(1) speech recognition using WhisperX~\cite{bain2022whisperx} resulting in punctuated transcriptions with word-level timestamps;
(2) sentence tokenization using nltk~\cite{bird2006nltk} to provide sentence-level segmentation;
(3) speaker diarization~\cite{Bredin2020,Bredin2021} to assign speaker labels to each sentence, where the sentence timestamps are used as oracle voice-activity-detection (VAD);
(4) labelling the speaker ID of the AD narrator by selecting the cluster with the lowest proportion of first-person pronouns (\eg~`I' and `we'); and finally
(5) synchronization of the segment timestamps with the visual features by comparing audio.
Further details are in the~\app.

Henceforth we refer to the original MAD annotation~\cite{soldan2022mad} as \textbf{MAD-v1} 
and our new denoised annotations as \textbf{MAD-v2}.
A qualitative comparison is shown in Fig.~\ref{fig:mad}, 
we find that our MAD-v2 is much more robust and contains less errors and less character dialogue leakage. 
Both LSMDC and MAD-v1 post-process their annotations by replacing character names 
in the annotations with `someone' via entity recognition, 
and release both variants of annotations which we refer to as \textbf{\texttt{Named}} and \textbf{\texttt{Unnamed}}. 
Similarly, we propose two variants of our denoised annotations: \\
\noindent\textbf{MAD-v2-\texttt{Named}:} It contains the raw collected AD narrations \emph{without} any post-processing on the character names. \\
\noindent\textbf{MAD-v2-\texttt{Unnamed}:} Following the character name anonymisation performed in earlier works, 
we identify character names using a Named Entity Recognition (NER) model~\cite{ner_jeanbaptiste} and replace them with `someone'.
\begin{table}
\centering
\scriptsize
\setlength{\tabcolsep}{2pt}
\resizebox{\columnwidth}{!}{%
\begin{tabular}{@{}lrrrcc@{}}
\toprule
Dataset                        & \begin{tabular}[c]{@{}l@{}}Total \\ movies\end{tabular} & \begin{tabular}[c]{@{}c@{}}Total\\ duration (hrs)\end{tabular} & \begin{tabular}[c]{@{}l@{}}Total AD\\ captions\end{tabular} & Subtitles & \begin{tabular}[c]{@{}l@{}}Visual\\ Features\end{tabular} \\ \midrule

QueryD~\cite{oncescu2021queryd} & - & 207 & 31K & \xmark & \cmark \\

LSMDC~\cite{rohrbach2017movie} & 200                                                     & 147                                                            & 128K                                                        & \xmark   & \cmark                                                    \\
MAD-v1~\cite{soldan2022mad}       & 488                                                     & 892                                                            & 280K                                                        & \cmark    & \cmark                                                    \\ \midrule
\textbf{MAD-v2 (ours)}                         & 488                                                     & 892                                                            & 264K                                                        & \cmark    & \cmark                                                    \\
\textbf{AudioVault (ours)}                     & 7,057                                                   & 12,510                                                         & 3.3M                                                        & \cmark    & \xmark                                                    \\ \bottomrule
\end{tabular}
}
\vspace{-2mm}
\caption{\textbf{Statistics of Audio Description datasets.} We report relevant statistics to compare our MAD-v2 and Audiovault datasets.}
\vspace{-5mm}
\label{tab:dataset-compare}
\end{table}

\vspace{-1mm}
\section{Partial Pretraining with AudioVault Dataset}
\label{sec:audiovault_textonly}
\vspace{-1mm}

Paired AD and corresponding visual data are difficult to obtain 
especially due to movie copyrights,
whereas a large number of movie ADs audio tracks are available online for free (\eg~AudioVault).
To demonstrate the effect of partial pretraining in Sec.~\ref{sec:ad-pretrain},
we collect a large-scale \emph{text-only} movie AD dataset from AudioVault.
In detail, we source mixed audio files from over 7,000 movies from AudioVault that are not included in MAD-v1,
and use a denoising pipeline similar to that described in Sec.~\ref{sec:mad}
to obtain the movie ADs (detailed in~\app).
Additionally we obtain a proxy for the movie subtitles by assuming the ASR from all the non-AD speakers are the characters' dialogues.
To ensure no test-time leakage, we remove all movies present in either LSMDC or MAD from the dataset. 

Overall, our AudioVault dataset is an order of magnitude larger 
than prior AD datasets (see Table~\ref{tab:dataset-compare}),
from which we provide two sets of data:

\noindent\textbf{AudioVault-AD.} 
The AD narrations from AudioVault and their corresponding timestamps within each movie, totalling 3.3 million AD utterances.

\noindent\textbf{AudioVault-Sub.} 
The subtitles data from AudioVault and their corresponding timestamps within each movie, totalling 8.7 million subtitle utterances.


\vspace{-2mm}
\section{Experiments}\vspace{-1mm}
\label{sec:exp}
In this section we first outline the experimental details for the AD task, the datasets used for training \& testing, the architectural details, and the evaluation metrics (Sec.~\ref{sec:impl}). We then report results and discuss the findings, 
perform ablations on our model, and compare to prior works (Sec.~\ref{exp:video}).

\vspace{-1mm}
\subsection{Implementation Details}
\label{sec:impl}

\subsubsection{Datasets}
\label{sec:impt:dataset}
\paragraph{Training Datasets.}
\textbf{CC3M}~(Conceptual Caption)~\cite{sharma2018cc}
is a large image alt-text dataset that contains 3.3M web images.
\textbf{WebVid}~\cite{Bain21} is
a large video-caption dataset that contains 2.5M short stock footage videos.
We use them for the partial-data pretraining for visual modules.
Additionally, we use our \textbf{AudioVault-AD} to pretrain the textual modules, 
as described in Sec.~\ref{sec:ad-pretrain}.
For the main Movie AD task, 
we train with original \textbf{MAD-v1} and our cleaned version \textbf{MAD-v2}, detailed in Sec.~\ref{sec:mad}. \\
\noindent\textbf{Test Datasets.}
\textbf{LSMDC}~\cite{rohrbach2015lsmdc}
contains 118K short video clips with descriptions from 202 movies, of which 182 of them are public.
The original {MAD}-val\&test split inherits LSMDC annotations 
after filtering out 20 lower-quality movies, resulting in 162 movies from all the LSMDC-train/val/test splits.
{
We propose an evaluation split named \textbf{MAD-eval} by further excluding LSMDC train\&test movies from these 162 movies, which gives a subset consisting of 10 movies.
The reason is twofold: 
(i) LSMDC-train is commonly used by other works as training data, and 
(ii) the character names of LSMDC-test are not public.
}
Similarly, we use both
\textbf{MAD-eval-\texttt{Named}} and \textbf{MAD-eval-\texttt{Unnamed}} versions.
The `Unnamed' version corresponds to the standard LSMDC annotation style -- where the characters' titles and names in the descriptions are replaced by the word `someone'; 
the `Named' version is constructed from the original character names provided by LSMDC.
Additionally, subtitles are not provided with MAD-val/test or LSMDC, 
so we transcribe them from the full-length audio tracks using WhisperX~\cite{bain2022whisperx}.

\vspace{-3mm}
\subsubsection{Architecture}\label{sec:exp:arch}

For \textbf{visual features}, 
we use the CLIP ViT-B-32 model~\cite{clip2021},
which is a 12-layer transformer encoder that outputs
$1\times512$ feature vectors for each input frame.
These features are provided by the MAD dataset.
For the \textbf{visual mapping network},
we use a 2-layer transformer encoder with 
8 attention heads and 512 hidden dimensions,
followed by a linear projection layer that projects $512$-d features into $768$-d.
We use ten prompt vectors.
For the \textbf{language model}, 
we use GPT-2~\cite{gpt2019}, specifically the version from HuggingFace.
The GPT-2 model takes as input $768$-d token embeddings,
passes through a 12-layer transformer with a causal attention map,
and outputs the next token embedding for every input token.
We limit the generated number of tokens to 36, since most movie ADs are less than 36 tokens.
The GPT-2 is frozen in most of our experiments unless otherwise stated.
Each special token (\eg~$\texttt{B}_{\text{AD}}$)
is a learnable $768$-d vector.
We take at most 64 past AD tokens and 32 subtitle tokens, and short text samples are padded.
Specifically for subtitles, we take the most recent four dialogues within a one-minute time window.

\vspace{-4mm}
\subsubsection{Training and Inference Details}
On the MAD-v1 and MAD-v2 datasets,
we use a batch size of 8 sequences,
each of which contains 16 consecutive video-AD pairs from a movie.
Overall that gives $8\times 16$ video-AD pairs for every batch.
From each video clip, 8 frame features are uniformly sampled.
By default, the model is trained for 10 epochs.
One epoch means the model has seen \emph{all} the audio descriptions once.
Additional implementation details are in the~\app.

We use the AdamW optimizer~\cite{loshchilov2017adamw} and a cosine-decay learning rate schedule with a linear warm-up.
The starting learning rate is $10^{-4}$ and is decayed to $0$.
For each experiment, we use a single Nvidia A-40 for training.
For text generation, greedy search and beam search are commonly used sampling methods. 
We stop the text generation when a full stop mark is predicted,
otherwise we limit the sequence length to 67 tokens.
We use beam search with a beam size of $5$
and mainly report results by the top-1 beam-searched outputs,
since beam search performs slightly better than greedy search on multiple scenarios.
Note that under the `recurrent' setting,
we feed the past greedy-searched text outputs to the model to generate the current AD,
which we find gives more stable results.

\vspace{-4mm}
\subsubsection{Evaluation Metrics}
To evaluate the quality of text compared with the ground-truth,
we use classic metrics including 
ROUGE-L~\cite{lin2004rouge}~(\textbf{R-L}), 
CIDEr~\cite{vedantam2015cider}~(\textbf{C}) and 
SPICE~\cite{anderson2016spice}~(\textbf{S}).
We also report BertScore~\cite{zhang2019bertscore}~(\textbf{BertS}),
which evaluates word matching between a candidate sentence and reference sentence with pretrained BERT embeddings.
A higher value indicates better text generation compared with the ground-truth.

\subsection{Experiments on Movie Audio Descriptions}
\label{exp:video}

\begin{table}
\centering
\scriptsize
\setlength{\tabcolsep}{3pt}
\resizebox{\columnwidth}{!}{%
\begin{tabular}{@{}ll|lllll@{}}
\toprule
\multirow{2}{*}{\begin{tabular}[c]{@{}l@{}}Temporal\\ Context\end{tabular}} & \multirow{2}{*}{\begin{tabular}[c]{@{}l@{}}Partial Data\\ Pretrain\end{tabular}} & \multirow{2}{*}{R-L} & \multirow{2}{*}{C} & \multicolumn{2}{l}{\multirow{2}{*}{S}} & \multirow{2}{*}{BertS} \\
                &                             &                      &                    & \multicolumn{2}{l}{}                   &                        \\ \midrule
None (1 frame)  & None                        & 7.1                  & 4.0                & 1.0                    &               & 13.2                   \\ \midrule
V (8 frames)    & None                        & 9.3                  & 6.7                & 2.4                    &               & 15.6                   \\
                & CC3M~\cite{sharma2018cc}    & 9.9                  & 8.4                & 2.4                    &               & 16.8                   \\
                & WV~\cite{Bain21}            & 9.9                  & 10.0               & 2.0                    &               & 17.3                   \\ \midrule
V+AD            & None                        & {11.1} (13.3)          & {12.6} (17.8)        & {5.1} (5.8)              &               & {18.6} (22.1)            \\
                & AV-AD                       & {12.1} (13.9)          & {14.1} (19.0)        & {4.2} (4.8)              &               & {23.0} (23.7)            \\
                & AV-AD, WV                   & {11.9} (13.9)          & {14.3} (21.9)        & {4.4} (4.8)              &               & {24.2} (23.8)            \\ \midrule
V+AD+Sub        & AV-AD, WV                   & {11.3}                 & {13.3}               & {4.7}                    &               & {22.2}                   \\
V+AD+SubN*      & AV-AD, WV                   & {11.9}                 & {14.2}               & {5.1}                    &               & {23.6}                   \\ \bottomrule
\end{tabular}
}
\vspace{-3mm}
    \caption{\textbf{Ablative experiments of our AD captioning method.} 
    We ablate our model with different types of temporal context and partial pretraining. 
    All models are trained on MAD-v2-\texttt{Named} and evaluated on MAD-eval-\texttt{Named}. For models with AD context we report recurrent results with oracle in parentheses.
    `V' refers to visual context by taking multi-frame inputs, `WV' refers to WebVid2M dataset, `AV-AD' here refers to our partial-data pretraining with text-only AudioVault-AD dataset.
    *`SubN' denotes the variant of subtitle module that only takes names as input.}
    \label{table:context}
    \vspace{-1mm}
\end{table}

\begin{table}
\scriptsize
\centering
\setlength{\tabcolsep}{4pt}
\resizebox{\columnwidth}{!}{%
\begin{tabular}{@{}cc|llll|llll@{}}
\toprule
\multicolumn{2}{c|}{\multirow{2}{*}{\begin{tabular}[c]{@{}c@{}}MAD\\ Train Set\end{tabular}}} & \multicolumn{4}{c}{MAD-eval-\texttt{Unnamed}}   & \multicolumn{4}{c}{MAD-eval-\texttt{Named}}  \\ \cmidrule(l){3-10} 
\multicolumn{2}{c|}{} & \multicolumn{1}{c}{R-L} & \multicolumn{1}{c}{C} & \multicolumn{1}{c}{S} & \multicolumn{1}{c}{BertS} & \multicolumn{1}{c}{R-L} & \multicolumn{1}{c}{C} & \multicolumn{1}{c}{S} & \multicolumn{1}{c}{BertS} \\ \midrule
\multirow{2}{*}{v1} & {\texttt{Unnamed}}  & 15.1 & 12.7 & 9.5  & 22.4        & {\color{gray}12.7} & {\color{gray}15.9} & {\color{gray}4.7}  & {\color{gray}22.0}   \\
                    & {\texttt{Named}}    & {\color{gray}11.3} & {\color{gray}10.9} & {\color{gray}3.0}  & {\color{gray}24.0}        & 12.8 & 17.0 & 5.2  & 21.8  \\ \midrule
\multirow{2}{*}{v2} & {\texttt{Unnamed}}  & \textbf{15.9} & \textbf{14.5} & \textbf{10.5} & \textbf{26.7}     & {\color{gray}12.9} & {\color{gray}{18.0}} & {\color{gray}4.7} & {\color{gray}22.0} \\
                    & {\texttt{Named}}    & {\color{gray}11.4} & {\color{gray}10.0} & {\color{gray}3.1}  & {\color{gray}22.5}      & \textbf{13.3} & \textbf{17.8}  & \textbf{5.8} & \textbf{22.1}     \\ \bottomrule
\end{tabular}
}
\vspace{-3mm}
\caption{\textbf{Effect of denoising MAD training data annotation.} 
We train a model with 6 contextual ADs on MAD-v1~\cite{soldan2022mad} or MAD-v2 sources without any pretraining.
The model is evaluated on both the \textbf{\texttt{Named}} and \textbf{\texttt{Unnamed}} versions of MAD-eval
under the \textbf{oracle} setting.
Cross-domain testing results (when the model is trained and tested on different types of annotations) 
are provided for reference and marked in {\color{gray}gray}.
}
\vspace{-1mm}
\label{table:cleaning}
\end{table}

\begin{table}

\centering
\scriptsize
\resizebox{\columnwidth}{!}{%
\begin{tabular}{@{}ll|llll@{}}
\toprule
Methods                         & Pretraining Data               & R-L & C   & S   & BertS \\ \midrule
ClipCap~\cite{mokady2021clipcap}  & CC3M   & 8.5 & 4.4 & 1.1 & 11.8  \\
CapDec*~\cite{nukrai2022text}     & AV-AD  & 8.2 & 6.7 & 1.4 & 14.3 \\
\midrule
AutoAD (ours)  & AV-AD            & {\textbf{12.1}} & {14.1}          & {4.2}  & {23.0} \\
AutoAD (ours)  & AV-AD \& WebVid  & {11.9}          & {\textbf{14.3}} & {\textbf{4.4}}           & \textbf{24.2} \\ \bottomrule
\end{tabular}
}
\vspace{-3mm}
\caption{Compared with other works on movie AD generation task on MAD-v2. 
We obtain results from other methods by finetuning their models 
on MAD-v2-\texttt{Named} dataset, and evaluated on MAD-eval-\texttt{Named}.
*CapDec~\cite{nukrai2022text} proposes text-only pretraining to adapt the style for text generation,
we pretrained their model on the text-only AudioVault-AD dataset 
then applied it to MAD-v2.
\label{tab:self}
\vspace{-2mm}
}

\end{table}
\begin{table}
\centering
\scriptsize
\resizebox{0.8\columnwidth}{!}{%
\begin{tabular}{@{}ll|ll@{}}
\toprule
Methods & \multicolumn{1}{c}{Paired Training Data} & \multicolumn{1}{|l}{C} & \multicolumn{1}{l}{M} \\ \midrule
Baseline~\cite{Park2019AdversarialIF} & LSMDC   & 11.9 & 8.3  \\
TAPM~\cite{yu2021transitional}        & LSMDC   & 15.4 & \textbf{8.4}  \\ \midrule
AutoAD (ours)  & MAD-v2-\texttt{Unnamed} & 16.7 & 7.4  \\
AutoAD (ours)  & MAD-v2-\texttt{Unnamed} \& LSMDC & \textbf{17.5} & 7.5 \\ 
\bottomrule
\end{tabular}
}
\vspace{-2mm}
\caption{
\textbf{Results on the LSMDC 2019 Multi-Sentence Description public test set.} We report our method with different amounts of training data and without subtitles for comparison under similar settings.
Official challenge metrics (CIDEr and METEOR) are reported with the `sentence' setting as described in~\cite{rohrbach2015lsmdc,yu2021transitional}. 
}
\vspace{-2mm}
\label{table:sota}
\end{table}

\begin{figure}
    \centering
    \includegraphics[width=0.45\textwidth]{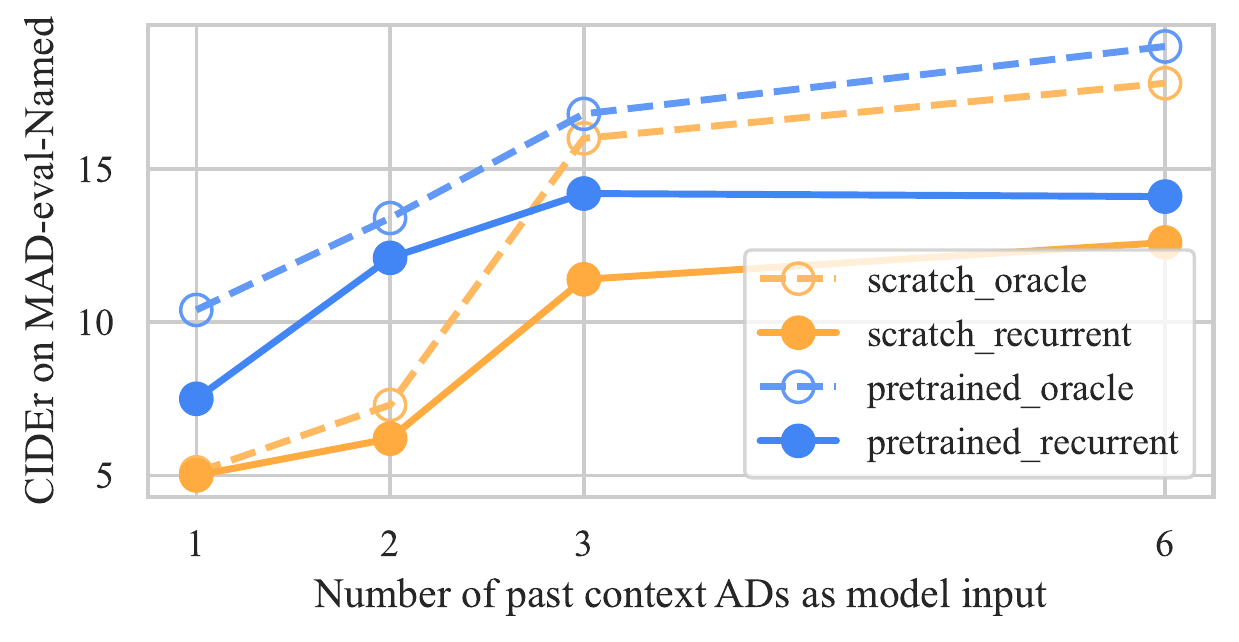}
    \vspace{-3mm}
    \caption{\textbf{Effect of the length of context AD.}
    We use the model `V+AD' in Table~\ref{table:context},
    and train with different number of past AD sentences. 
    `scratch' indicates no partial-data pretraining; `pretrained' refers to pretraining with text-only AudioVault-AD.
    }
    \label{fig:length}
    \vspace{-3mm}
\end{figure}

\begin{figure*}[h!]
    \centering
    \includegraphics[width=0.98\textwidth]{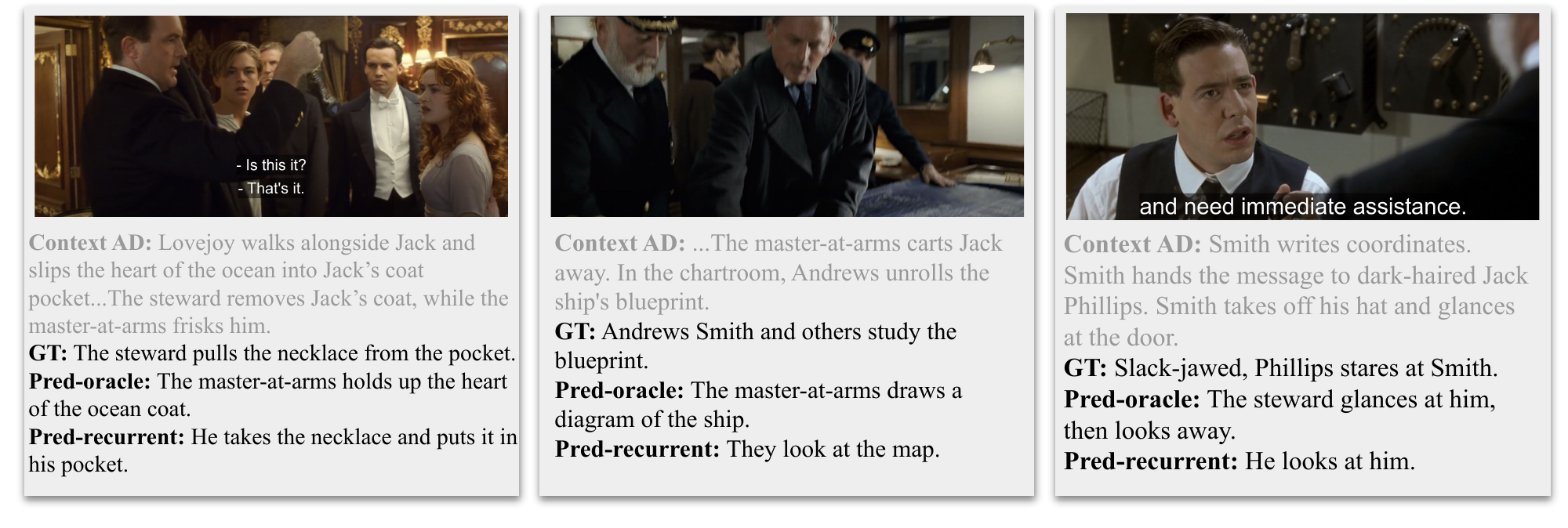}
    \vspace{-5mm}
    \caption{\textbf{Qualitative examples of automatically generated AD by AutoAD.}
    We highlight AD predictions under both the oracle and recurrent settings.
    Previous AD context is shown in {\color{gray}gray}. For ease of visualisation, a single frame from each movie clip is shown with subtitles overlaid. 
    Samples are taken from Titanic (1997)~\cite{titanic}.
    }
    \vspace{-5mm}
    \label{fig:mad}
\end{figure*}

\paragraph{Effect of Temporal Context.}
In Table~\ref{table:context} we show that visual context from multiple frames brings a clear gain for the AD task (C 6.7 vs 4.0).
AD context provides a consistent performance improvement under both oracle (C 17.8 vs 6.7) and recurrent settings (C 12.6 vs 6.7).
Note that we find feeding AD context as text tokens works better than
training a textual feature mapping network, we conjecture the ADs in their original text form carry the most key information like the names and places.
However, subtitle context provides no gain for our model (C 13.3 vs 14.3) under the recurrent setting, 
which we attribute to the very weak correspondence between the visual elements in the scene and the character dialogue. 
When the subtitles are filtered and contain only character names (denoted as `SubN'),
they provide a slight performance gain (C 14.2 vs 13.3).
Since the subtitles used are without speaker identities, the model may struggle to know which character in the frame spoke each subtitle. Overcoming these challenges will be considered in future work.

\vspace{-2mm}
\paragraph{Effect of MAD data cleaning.}
Table~\ref{table:cleaning} demonstrates the benefit of our  MAD v2 annotations over v1, 
confirming the qualitative findings. 
Training the AD model with context on v2 outperforms training on v1 under all settings (both named and unnamed) by a significant margin.
Since the v2 annotations are fewer in number than MAD-v1, 
this suggests they are indeed less noisy and result in AD captioning models with improved performance. 
\vspace{-2mm}
\paragraph{Effect of Pretraining with Partial Data.}
In Table~\ref{table:context},
we find that \textbf{visual-only pretraining} on open-domain vision-text data provides clear gains 
(CIDEr 8.4 vs 6.7 for CC3M, and 10.0 vs 6.7 for WebVid).
But considering the size of visual samples, the improvement is not data-efficient.
We attribute this to the large domain gap between movie AD and classical visual caption annotations like CC3M or WebVid2M.
{
The \textbf{text-only pretraining} of our model also improves performance. 
For the recurrent AD context model, 
AudioVault-AD pretraining increases CIDEr from 12.6 to 14.1, which indicates the great importance of adapting to the text style and context.}
The combination of the visual module after visual-only pretraining (WebVid) and the textual modules after text-only pretraining (AV-AD) gives a further performance gain 
(C 21.9 vs 19.0 for the oracle setting, and 14.3 vs 14.1 for recurrent). \\
\vspace{-8mm}
\paragraph{Length of Context.}
{In Figure~\ref{fig:length} we show the effect of varying the number of context ADs given to the model. 
Longer AD context improves performance almost consistently across all settings, but it brings extra computational cost due to the quadratic complexity of the attention operation in GPT-2. Note that we experiment with at most 6 contextual AD sentences, which is equivalent to about 70-word embeddings in Eq.~\ref{eq:context_ad}.
The trend for the recurrent setting flattens when the context ADs are longer than 3 sentences, 
which is probably due to the limited power of processing long context for the GPT2 model.
}

\vspace{-3mm}
\subsubsection{Qualitative Results}
Fig.~\ref{fig:mad} shows qualitative examples of our model.
Under the oracle setting, the model can use the character identities easily from the past ground-truth AD (\eg~``master-at-arms'').
Whereas under the recurrent setting,
the model can only learn names from the subtitles but names appear very sparsely in subtitles,
therefore the model mostly predicts pronouns (\eg~``he'', ``they'') but still gets the actions (``looks'') or objects (``necklace'') correct.

\subsection{Comparison with Other Works}
In Table~\ref{tab:self},
we compare our method with previous visual captioning methods. Note that since the MAD dataset only releases the CLIP visual features, rather than the movie frames,
our comparison is limited to methods that build on frozen CLIP features.
We show a clear performance improvement compared to ClipCap~\cite{mokady2021clipcap} and CapDec~\cite{nukrai2022text}, for the latter the language model is also adapted to the movie AD domain by text-only pretraining. The results highlight the importance of context for movie AD.

In Table~\ref{table:sota},
we adapt our method to the Multi-Sentence Description task on LSMDC, in which the model takes five consecutive clips and generates five corresponding descriptions. Since the task is performed on the \emph{unnamed} annotations, we finetune our best model in Table~\ref{tab:self} 
with varying 0-4 context ADs as input on MAD-v2-\texttt{Unnamed} dataset and test with the \emph{recurrent} setting.
To make minimal changes,
our model still takes a single clip feature at each step,
whereas previous methods take all five clips together for movie description.
Despite this disadvantage, we obtain competitive results on this task
even without using the \emph{manually-cleaned} LSMDC training set (C 16.7 vs 15.4), effectively \emph{zero-shot}.
The performance of the model can be further improved by additionally training on LSMDC data.
\section{Conclusion and Future Work}

This paper focuses on the automatic generation of movie AD for a given time interval, and has made significant progress.
We propose an AutoAD pipeline that incorporates contextual information. Additionally, we demonstrate the effectiveness of partial-data pretraining, a technique that could be widely applicable when full data is difficult to obtain.
Further, we clean up the previous MAD dataset and collect a new text-only movie AD dataset as a pretraining resource.
{
However, a clear limitation of this AutoAD pipeline is character naming -- referencing \textit{who} is doing \textit{what}, a necessary ingredient for story-coherent movie AD. Additionally,
future work could tackle the problem of \textit{when} to generate AD,
instead of relying on the annotated AD timestamps.
}
\label{sec:conclusion}

\vspace{2mm}
{
\noindent\textbf{Acknowledgements.}
We thank Mattia Soldan for helping with the MAD dataset,
Anna Rohrbach for the LSMDC dataset,
and the AudioVault team for their priceless contribution to the visually impaired.
This research is funded by EPSRC PG VisualAI EP/T028572/1,
a Google-Deepmind Scholarship, and
ANR-21-CE23-0003-01 CorVis.
\par
}
\clearpage

{\small
\bibliographystyle{ieee_fullname}
\bibliography{bib/shortstrings,bib/vgg_local,bib/vgg_other}
}

\clearpage
\onecolumn
\appendix
\noindent{\Large{\textbf{Appendix}}}
\vspace{5mm}
\renewcommand{\thefigure}{A.\arabic{figure}} 
\setcounter{figure}{0} 
\renewcommand{\thetable}{A.\arabic{table}}
\setcounter{table}{0} 

\appendix

We first show the details of the AD collection pipeline (Sec.~\ref{app:sec:details})
with qualitative text examples (Sec.~\ref{app:sec:qualitative-text}).
Then we describe additional implementation details (Sec.~\ref{app:sec:implementation})
with extra qualitative movie AD examples (Sec.~\ref{app:sec:qualitative-ad}).
Finally, we list the movie IDs used in our MAD-v2 split (Sec.~\ref{app:sec:ids}).

\section{AD Collection Pipeline Additional Details }
\label{app:sec:details}

\begin{figure}[h!]
    \centering
    \includegraphics[width=\textwidth]{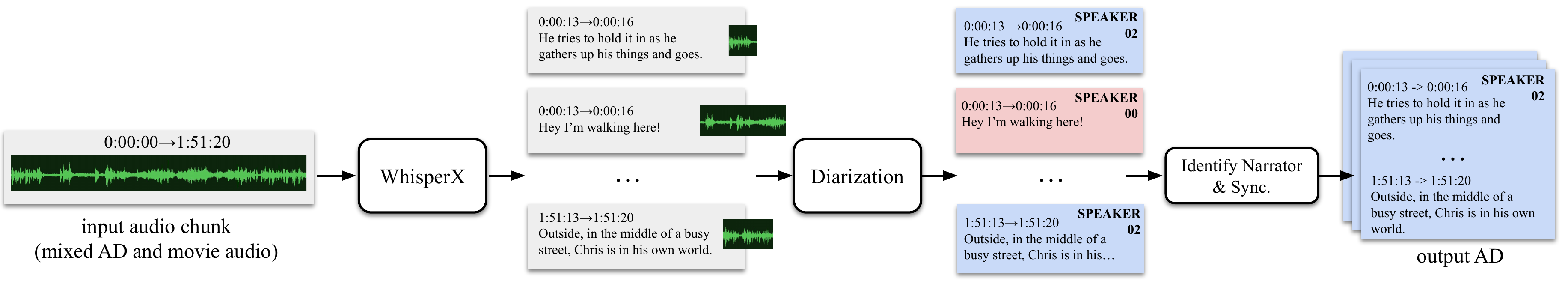}
    \caption{\textbf{A schematic of our AD collection pipeline.}
    The pipeline takes the audio file (with mixed AD and movie audio) as input,
    and automatically outputs the AD in text form with corresponding timestamps. 
    }
    \label{fig:pipeline}
\end{figure}

\subsection{AD Collection Pipeline for MAD-v2}
Collecting movie AD has two main challenges.
First, in the audio files (\eg~from AudioVault)
the movie AD is \emph{fused} with the original movie audio,~\ie~on the same audio track. The pipeline needs to identify the AD speaker among the movie characters accurately.
Second, for the same movie,
the audio files from AudioVault is usually not synchronised with the movie from which the MAD visual features were extracted,
mainly due to the varied durations of intro and outro of different movie source.
Since we rely on the MAD visual features, the synchronisation is an essential step.

The automated data collection pipeline is briefly introduced in Sect.~\ref{sec:mad} of the main paper.
A schematic is shown in Fig.~\ref{fig:pipeline}, in detail:

\begin{enumerate}
\item We transcribe the mixed audio file using \textit{WhisperX}~\cite{bain2022whisperx} which provides accurate punctuated transcriptions with word-level timestamps.
\item The transcript is tokenized into sentences using the nltk python toolbox~\cite{bird2006nltk},  resulting in transcription sentences and their corresponding temporal segments (inferred the start and end time of the first and last word in the sentence respectively).
\item Each sentence segment is assigned a single speaker identity (\eg~\texttt{SPEAKER\_00}, \texttt{SPEAKER\_01}, \etc) by performing speaker diarization on the mixed audio, whereby each sentence timestamp is provided as oracle voice activity detection. Specifically,  we use SpeechBrain ECAPA-TDNN voice embeddings~\cite{desplanques2020ecapa} trained on VoxCeleb~\cite{nagrani2017voxceleb} and Agglomerative Clustering with a threshold of 0.95. \item To automatically identify the cluster associated with the AD speaker, we exploit the third-person nature of AD narrations and select the cluster with the lowest proportional occurrence of first- \& second-person pronouns,~\eg~``I'' and ``you'' with 95 or more speaker segments.
\item To synchronise the segment timestamps with the original audio track from which the MAD visual features were extracted, we follow~\cite{soldan2022mad} and calculate the time delay $\tau$ between the original movie audio files and the mixed audio files via FFT cross-correlation. The timestamps of the identified AD segments are shifted according $\tau$ in order to synchronise them to the visual features and subtitles collected in MAD.
\end{enumerate}

\subsection{AD Collection Pipeline for AudioVault}

The collection pipeline for AudioVault is introduced in  Sect.~\ref{sec:audiovault_textonly} of the main paper, 
we provide more details here.
To collect text-only AD annotations from AudioVault, the final synchronisation step is unnecessary.
Therefore, we follow steps 1-4 of the MAD denoising pipeline as described above,
which takes as input the mixed audio tracks and
outputs the ASR with timestamps from the possible AD speaker.

The large-scale collection from AudioVault audio files is noisy,
\eg~some ADs are of lower-quality or are sourced from short movies. Therefore, we apply a stricter filtering step that 
removes movies containing fewer than 100 AD narrations 
or a word frequency of first- \& second-person pronouns larger than 5\%.

\subsection{Comparison with MAD-v1}
The key advantages of our pipeline are three-fold:
(1) it relies on \textit{audio-based} speaker separation to identify the AD speaker among the movie characters, 
whereas the pipeline in the original MAD work~\cite{soldan2022mad} relies on \textit{text-based} speaker separation by using the timestamps from the DVD subtitles and assumes any ASR transcription outside of these timestamps is AD. The error is propagated because the official subtitles are non-exhaustive (some dialogue is missed by the official subtitles). (2) It requires only the mixed audio as input, whereas MAD must also source the official DVD subtitles and align them -- presenting additional scaling costs and challenges. (3) It uses an advanced ASR model Whisper~\cite{radford2022whisper} which gives much more accurate transcriptions than previous methods, especially for punctuation and the spelling of names and other identities.

\section{Qualitative Examples of MAD-v2 vs MAD-v1.}
\label{app:sec:qualitative-text}
\begin{figure}[h!]
    \centering
    \includegraphics[width=\textwidth]{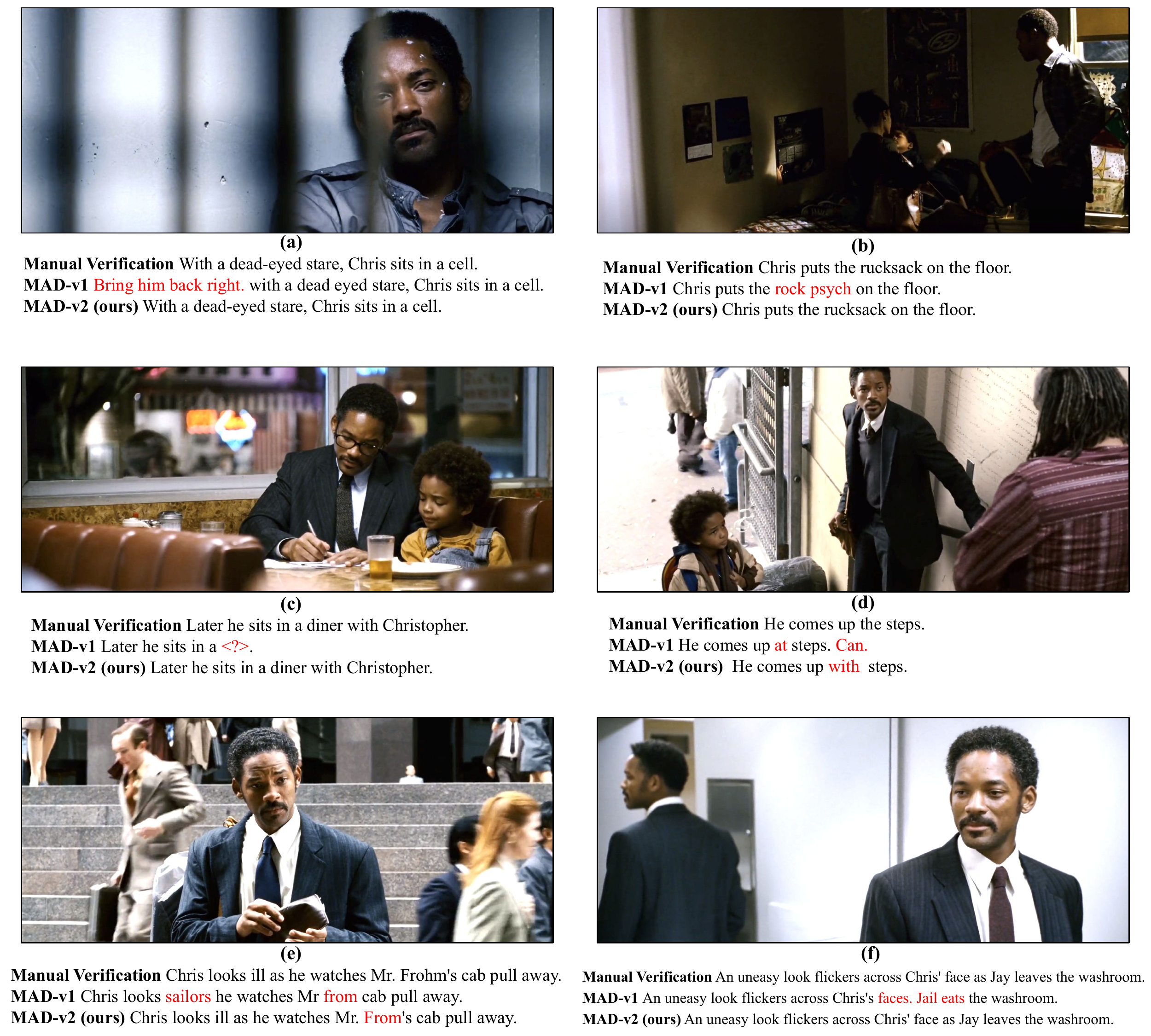}
    \caption{\textbf{Comparison of the AD quality from MAD-v2 with MAD-v1.}
    The erroneous transcriptions are marked in {\color{red} red text}. 
    `Manual Verification' means we manually transcribe the AD narration from the audio track.
    The sample is originally from \emph{The Pursuit of Happyness} (2006).
    The failure mode of MAD-v1 in each example is 
    \textbf{(a)} dialogue leakage, \textbf{(b)} incorrect ASR, \textbf{(c)} missing words,
    \textbf{(d)} dialogue leakage, \textbf{(e)} incorrect ASR and name spelling,
    \textbf{(f)} incorrect name spelling.
    }
    \label{fig:happyness}
\end{figure}

\begin{figure}[h!]
    \centering
    \includegraphics[width=\textwidth]{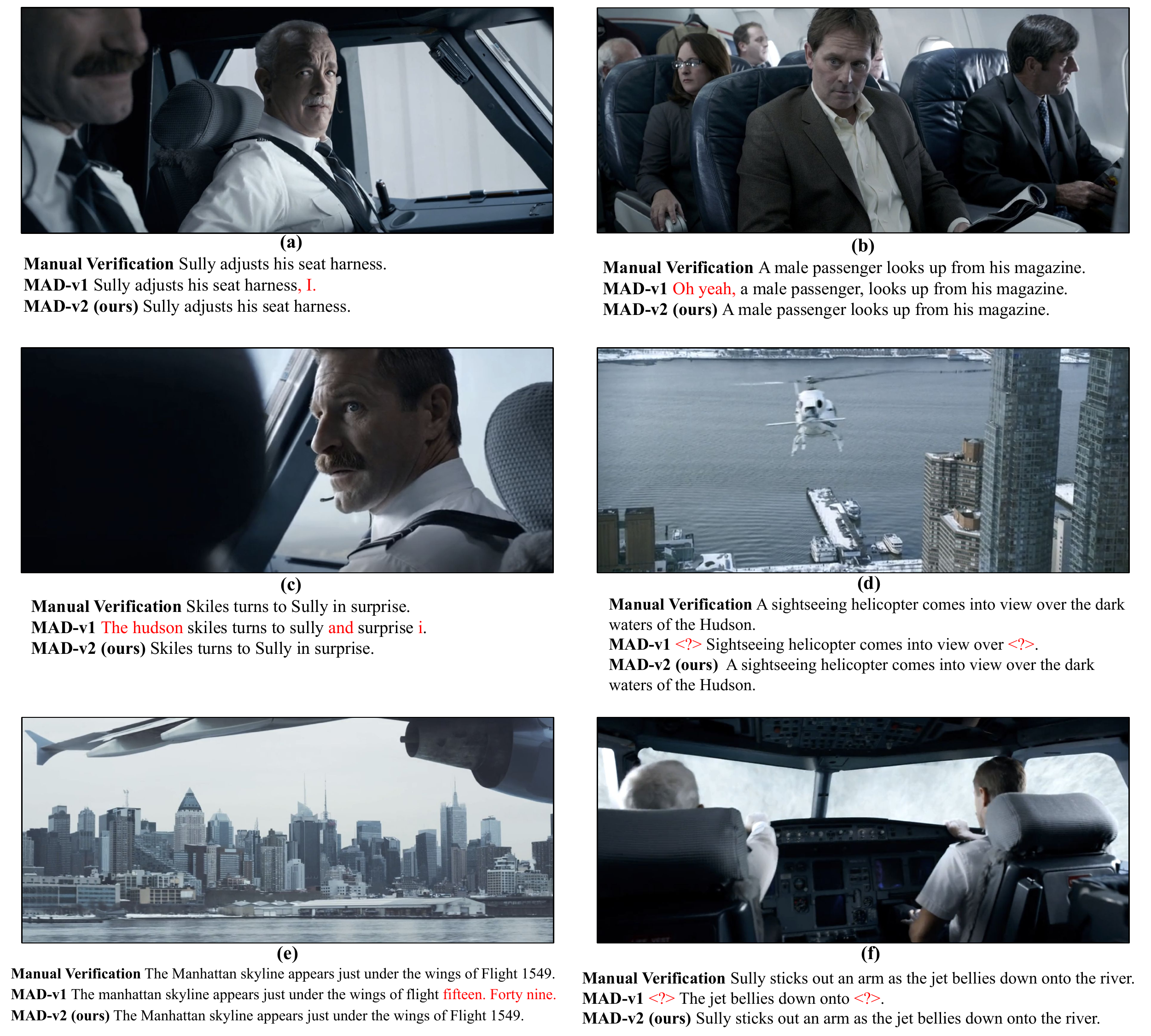}
    \caption{\textbf{(continue) Comparison of the AD quality from MAD-v2 with MAD-v1.}
    The erroneous transcriptions are marked in {\color{red} red text}. 
    `Manual Verification' means we manually transcribe the AD narration from the audio track.
    The sample is originally from \emph{Sully: Miracle on the Hudson} (2016).
    The failure mode of MAD-v1 in each example is 
    \textbf{(a)} dialogue leakage, \textbf{(b)} dialogue leakage, 
    \textbf{(c)} dialogue leakage and incorrect ASR,
    \textbf{(d)} missing words, \textbf{(e)} number spelling and sentence partitioning,
    \textbf{(f)} missing words.
    }
    \label{fig:sully}
\end{figure}

More qualitative examples 
of MAD-v2 and MAD-v1 are shown in Fig.~\ref{fig:happyness} and~\ref{fig:sully}.
It is clear that our pipeline produces more accurate AD
compared to the original MAD-v1,
particularly in the spelling of names and 
the exclusion of dialogue.

\clearpage
{
\section{Quantitative Comparison between MAD-v2 vs MAD-v1 on Grounding}
{
We re-purpose the CLIP zero-shot video-language grounding (VLG) performance from~\cite{soldan2022mad} as an indicator of dataset quality.
In detail, for both MAD-v2 and MAD-v1, 
we randomly choose a set of 5 movies from the \emph{training split},
and compute the VLG performance with frozen CLIP visual and textual encoders.
The AD textual quality and timestamps are the only factors that differ in this comparison.
We use the MAD training split because we did not modify the val/test splits, 
which are from LSMDC annotations.
The code to compute VLG performance is from~{\small\url{https://github.com/Soldelli/MAD}}.
The result in Table~\ref{tab:vlg} shows MAD-v2 annotations also benefit the VLG task.
}
\vspace{-2mm}
\begin{table}[h]
\centering
\small
\begin{tabular}{l|lll}
\hline
R@50 & IoU@0.1 & IoU@0.3 & IoU@0.5 \\ \hline
MAD-v1-\texttt{Unnamed} & 32.08 & 22.85 & 14.26 \\
MAD-v2-\texttt{Unnamed} & \textbf{33.25} & \textbf{24.22} & \textbf{15.58} \\ \hline
\end{tabular}
\vspace{-3mm}
\caption{CLIP zero-shot VLG performance on MAD-v1 and MAD-v2.}
\label{tab:vlg}
\vspace{-2mm}
\end{table}
}

\section{Additional Implementation Details}
\label{app:sec:implementation}
{
\paragraph{Design Choices.}
\begin{itemize}
\setlength\itemsep{-1mm}
    \item Number of frames per movie clip $N$: We choose $N=8$. Most AD annotations have a time duration of 1-3 seconds, equivalent to 5-15 frames (features) under 5 FPS -- the sampling rate provided by the MAD dataset. Therefore $N=8$ is a reasonable choice.
    \item Number of AD sentences as context $K$: We experiment $K\in\{1,2,3,6\}$ in Figure~\ref{fig:length}.
    \item Number of subtitles $L$: As described in Sect.~\ref{sec:exp:arch}, for simplicity we take the most recent 4 dialogues within a 1-minute time window. Note that the time distribution of subtitles varies a lot -- the most recent 4 dialogues could span just a few seconds or up to minutes before the current AD timestamp.
\end{itemize}
}

\paragraph{Evaluation Metrics.}
We use the \texttt{pycocoeval} package from~{\small \url{https://github.com/tylin/coco-caption}}
to compute the ROUGE-L, CIDEr, SPICE and METEOR.
The package post-processes both the predicted text and ground-truth text internally to remove the punctuation and make them lowercase.
To compute the BertScore, 
we use the package from~{\small \url{https://github.com/Tiiiger/bert_score}}.
Note that before computing the BertScore, 
both the predicted text and the ground-truth text are 
converted to lowercase without any punctuation,
as these are factors that the BertScore is sensitive to.

{
\paragraph{Alternative approach for vision-language fusion.}
We investigate an alternative vision \& language fusion mechanism whereby the context AD sentence prompts are fed as \emph{language features} rather than \emph{raw text
tokens}. 
Empirically, we observe that raw text inputs outperform language features 
(e.g. 12.6 CIDEr in Table~\ref{table:context} vs. about 8.0 CIDEr when feeding language features).
}

\section{Additional Qualitative Examples}
\label{app:sec:qualitative-ad}

\begin{figure}[t!]
    \centering
    \includegraphics[width=\textwidth]{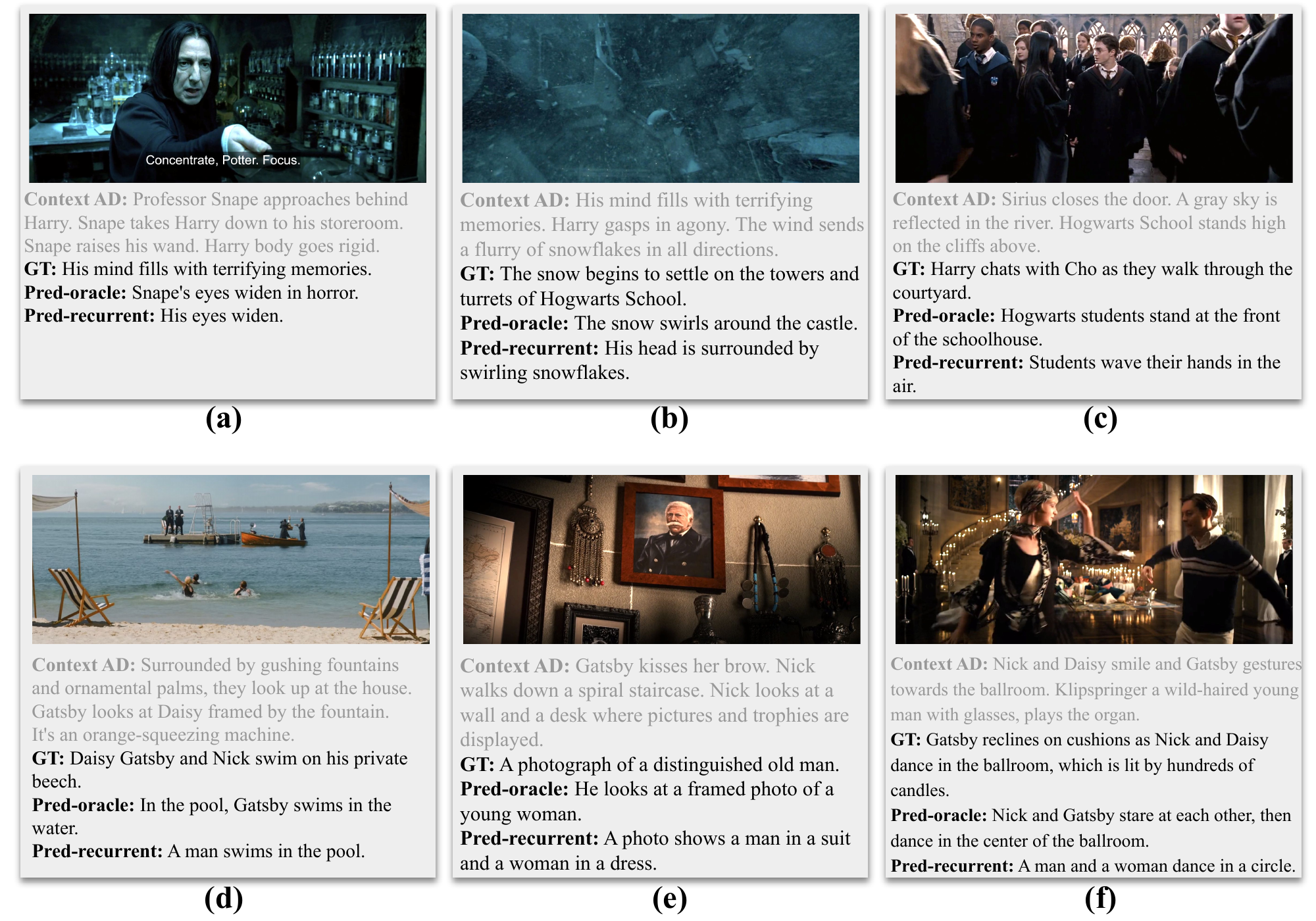}
    \caption{\textbf{Qualitative examples of AutoAD model.}
    We show the ground-truth AD and the AD predictions 
    under both the oracle and recurrent settings. 
    Previous AD context is shown in {\color{gray}gray}.
    Samples are taken from \emph{Harry Potter and the Order of the Phoenix} (2007) and \emph{The Great Gatsby} (2013).
    }
    \label{fig:more_qualitative}
\end{figure}

More qualitative examples are shown in Fig.~\ref{fig:more_qualitative}.
It shows that the AutoAD model gives reasonable descriptions
for the movie domain, 
like the actions (swim, dance), and face expression (eyes widen).
Note that under the oracle setting, the model is capable of learning character names (sample \textbf{a, c, d, f}) mainly due to the extra information from the ground-truth context. 
The model is still limited in its ability to identify characters accurately,~\eg in sample \textbf{f}, the movie shows Nick and \emph{Daisy} are dancing. Whereas the oracle prediction describes that Nick and \emph{Gatsby} are dancing, and the recurrent model simply predicts that a man and woman are dancing. {Also the pronouns often appear in the recurrent prediction, such as the word `his' in sample \textbf{a} and \textbf{b}, which shows the model learns the bias of pronouns but cannot recognize characters correctly.}
\section{Dataset Splits}
\label{app:sec:ids}

To clarify the dataset split of the movies in MAD and LSMDC, 
we list the movie IDs of each split used (and not used) in this paper.
The splits can also be found on the website~\small{\url{https://www.robots.ox.ac.uk/~vgg/research/autoad/}}.

\paragraph{MAD-v2.} 
It consists of 488 movies and all of them are used for training.
We provide the cleaner ADs for these movies using the automated pipeline described above.
This set is the same as the training set of movies proposed in MAD~\cite{soldan2022mad}. 
The movie IDs are:\\
{\scriptsize\texttt{
[2723, 2730, 2731, 2735, 2738, 2745, 2750, 2758, 2768, 2778, 2787, 2800, 2801, 2814, 2818, 2854, 2869, 2870, 2873, 2911, 2913, 2928, 2934, 2944, 2948, 2970, 2986, 2992, 2996, 3001, 3014, 3020, 3021, 3023, 3033, 3040, 3049, 3050, 3059, 3060, 3066, 3070, 3103, 3106, 3113, 3114, 3117, 3129, 3138, 3146, 3153, 3160, 3170, 3171, 3209, 3239, 3253, 3276, 3277, 3295, 3314, 3339, 3340, 3354, 3376, 3393, 3401, 3408, 3414, 3417, 3447, 3464, 3480, 3482, 3500, 3509, 3510, 3513, 3521, 3548, 3575, 3590, 3599, 3611, 3625, 3720, 3743, 3759, 3773, 3820, 3834, 3837, 3858, 3905, 3911, 3922, 3977, 4001, 4007, 4010, 4017, 4031, 4043, 4053, 4061, 4071, 4080, 4082, 4143, 4156, 4200, 4204, 4210, 4253, 4266, 4299, 4303, 4305, 4368, 4377, 4378, 4390, 4423, 4434, 4451, 4455, 4460, 4480, 4489, 4528, 4535, 4551, 4576, 4578, 4587, 4596, 4597, 4608, 4611, 4618, 4634, 4635, 4638, 4644, 4664, 4670, 4671, 4684, 4702, 4709, 4719, 4728, 4740, 4741, 4753, 4772, 4778, 4797, 4798, 4813, 4815, 4839, 4880, 4884, 4888, 4901, 4902, 4914, 4925, 4929, 4933, 4936, 4950, 4962, 4970, 4977, 4982, 4992, 5014, 5041, 5055, 5063, 5074, 5093, 5101, 5118, 5139, 5144, 5217, 5236, 5237, 5257, 5259, 5265, 5270, 5283, 5293, 5308, 5335, 5366, 5367, 5369, 5417, 5420, 5432, 5449, 5461, 5469, 5473, 5477, 5494, 5506, 5510, 5511, 5522, 5563, 5565, 5568, 5574, 5575, 5577, 5583, 5594, 5605, 5607, 5634, 5641, 5649, 5677, 5678, 5682, 5685, 5700, 5735, 5737, 5743, 5749, 5752, 5758, 5762, 5792, 5807, 5814, 5818, 5819, 5828, 5852, 5865, 5872, 5873, 5898, 5900, 5913, 5923, 5950, 5958, 6012, 6013, 6022, 6048, 6055, 6057, 6076, 6086, 6090, 6137, 6153, 6154, 6156, 6177, 6186, 6194, 6224, 6232, 6319, 6334, 6394, 6402, 6491, 6521, 6607, 6613, 6617, 6629, 6636, 6655, 6656, 6672, 6685, 6701, 6706, 6741, 6769, 6770, 6775, 6810, 6811, 6816, 6819, 6832, 6833, 6837, 6859, 6869, 6870, 6878, 6890, 6952, 6959, 6992, 6994, 7001, 7005, 7007, 7026, 7036, 7050, 7055, 7131, 7195, 7196, 7243, 7682, 7882, 8152, 8276, 8295, 8346, 8496, 8578, 8587, 8589, 8593, 8598, 8601, 8608, 8616, 8618, 8637, 8734, 8766, 8767, 8811, 9110, 9277, 9380, 9384, 9386, 9387, 9419, 9421, 9451, 9456, 9460, 9461, 9462, 9481, 9482, 9488, 9502, 9504, 9509, 9510, 9515, 9519, 9526, 9528, 9529, 9535, 9552, 9555, 9575, 9576, 9583, 9595, 9606, 9615, 9617, 9618, 9619, 9620, 9638, 9642, 9644, 9647, 9654, 9659, 9676, 9689, 9719, 9724, 9732, 9733, 9735, 9737, 9738, 9741, 9747, 9750, 9751, 9754, 9756, 9761, 9773, 9774, 9785, 9799, 9846, 9896, 9906, 9920, 9952, 10142, 10149, 10202, 10322, 10527, 10536, 10784, 10813, 10836, 10861, 10894, 10965, 11003, 11010, 11099, 11129, 11139, 11140, 11143, 11147, 11148, 11154, 11318, 11321, 11345, 11396, 11430, 11438, 11530, 11620, 11727, 11796, 11962, 12010, 12079, 12090, 12125, 12131, 12132, 12144, 12147, 12148, 12186, 12211, 12220, 12222, 12263, 12273, 12294, 12324, 12358, 12504, 12563, 12585, 12618, 12653, 12658, 12743, 12852, 12869, 12900, 12906, 12911, 12923, 12958, 13018, 13027, 13031, 13045, 13140, 13146, 13159, 13165, 13187, 13191, 13201]
}}

\paragraph{MAD-eval Evaluation Set.}
{
It consists of 10 movies,
which are obtained by 
$ set(\text{MAD val/test}) \cap set(\text{LSMDC val}) $,
that excluding LSMDC train movies for the ease of future comparison.
The annotations are inherited from the LSMDC dataset,
and we use both the \emph{named} and \emph{unnamed} version of it,
where the \emph{named} version can be downloaded from the LSMDC website\footnote{https://sites.google.com/site/describingmovies/download?authuser=0}.
The movie IDs are:\\
{\scriptsize\texttt{
[1005\_Signs,
1026\_Legion,
1027\_Les\_Miserables,
1051\_Harry\_Potter\_and\_the\_goblet\_of\_fire,
3009\_BATTLE\_LOS\_ANGELES,
3015\_CHARLIE\_ST\_CLOUD,
3031\_HANSEL\_GRETEL\_WITCH\_HUNTERS,
3032\_HOW\_DO\_YOU\_KNOW,
3034\_IDES\_OF\_MARCH,
3074\_THE\_ROOMMATE]}}
}

{
\paragraph{Unused MAD movies.}
151 movies from MAD val/test are \emph{not used} in either training or testing in our paper.
They are the intersection $ set(\text{MAD val/test}) \cap set(\text{LSMDC train/test}) $.
The movie IDs are:\\
{\scriptsize\texttt{
[0001\_American\_Beauty, 0002\_As\_Good\_As\_It\_Gets, 0003\_CASABLANCA, 0004\_Charade, 0005\_Chinatown, 0006\_Clerks, 0007\_DIE\_NACHT\_DES\_JAEGERS, 0008\_Fargo, 0009\_Forrest\_Gump, 0010\_Frau\_Ohne\_Gewissen, 0011\_Gandhi, 0012\_Get\_Shorty, 0013\_Halloween, 0014\_Ist\_das\_Leben\_nicht\_schoen, 0016\_O\_Brother\_Where\_Art\_Thou, 0017\_Pianist, 0019\_Pulp\_Fiction, 0020\_Raising\_Arizona, 0021\_Rear\_Window, 0022\_Reservoir\_Dogs, 0023\_THE\_BUTTERFLY\_EFFECT, 0026\_The\_Big\_Fish, 0027\_The\_Big\_Lebowski, 0028\_The\_Crying\_Game, 0029\_The\_Graduate, 0030\_The\_Hustler, 0031\_The\_Lost\_Weekend, 0032\_The\_Princess\_Bride, 0033\_Amadeus, 0038\_Psycho, 0041\_The\_Sixth\_Sense, 0043\_Thelma\_and\_Luise, 0046\_Chasing\_Amy, 0049\_Hannah\_and\_her\_sisters, 0050\_Indiana\_Jones\_and\_the\_last\_crusade, 0051\_Men\_in\_black, 0053\_Rendezvous\_mit\_Joe\_Black, 1001\_Flight, 1002\_Harry\_Potter\_and\_the\_Half-Blood\_Prince, 1003\_How\_to\_Lose\_Friends\_and\_Alienate\_People, 1004\_Juno, 1006\_Slumdog\_Millionaire, 1007\_Spider-Man1, 1008\_Spider-Man2, 1009\_Spider-Man3, 1010\_TITANIC, 1011\_The\_Help, 1012\_Unbreakable, 1014\_2012, 1015\_27\_Dresses, 1017\_Bad\_Santa, 1018\_Body\_Of\_Lies, 1019\_Confessions\_Of\_A\_Shopaholic, 1020\_Crazy\_Stupid\_Love, 1028\_No\_Reservations, 1031\_Quantum\_of\_Solace, 1033\_Sherlock\_Holmes\_A\_Game\_of\_Shadows, 1034\_Super\_8, 1035\_The\_Adjustment\_Bureau, 1037\_The\_Curious\_Case\_Of\_Benjamin\_Button, 1038\_The\_Great\_Gatsby, 1039\_The\_Queen, 1040\_The\_Ugly\_Truth, 1042\_Up\_In\_The\_Air, 1043\_Vantage\_Point, 1045\_An\_education, 1046\_Australia, 1047\_Defiance, 1048\_Gran\_Torino, 1050\_Harry\_Potter\_and\_the\_deathly\_hallows\_Disk\_One, 1052\_Harry\_Potter\_and\_the\_order\_of\_phoenix, 1054\_Harry\_Potter\_and\_the\_prisoner\_of\_azkaban, 1055\_Marley\_and\_me, 1057\_Seven\_pounds, 1058\_The\_Damned\_united, 1059\_The\_devil\_wears\_prada, 1060\_Yes\_man, 1061\_Harry\_Potter\_and\_the\_deathly\_hallows\_Disk\_Two, 1062\_Day\_the\_Earth\_stood\_still, 3001\_21\_JUMP\_STREET, 3002\_30\_MINUTES\_OR\_LESS, 3003\_40\_YEAR\_OLD\_VIRGIN, 3004\_500\_DAYS\_OF\_SUMMER, 3005\_ABRAHAM\_LINCOLN\_VAMPIRE\_HUNTER, 3007\_A\_THOUSAND\_WORDS, 3008\_BAD\_TEACHER, 3012\_BRUNO, 3013\_BURLESQUE, 3014\_CAPTAIN\_AMERICA, 3016\_CHASING\_MAVERICKS, 3017\_CHRONICLE, 3018\_CINDERELLA\_MAN, 3020\_DEAR\_JOHN, 3022\_DINNER\_FOR\_SCHMUCKS, 3023\_DISTRICT\_9, 3024\_EASY\_A, 3025\_FLIGHT, 3026\_FRIENDS\_WITH\_BENEFITS, 3028\_GHOST\_RIDER\_SPIRIT\_OF\_VENGEANCE, 3030\_GROWN\_UPS, 3033\_HUGO, 3035\_INSIDE\_MAN, 3036\_IN\_TIME, 3037\_IRON\_MAN2, 3038\_ITS\_COMPLICATED, 3039\_JACK\_AND\_JILL, 3040\_JULIE\_AND\_JULIA, 3041\_JUST\_GO\_WITH\_IT, 3042\_KARATE\_KID, 3043\_KATY\_PERRY\_PART\_OF\_ME, 3045\_LAND\_OF\_THE\_LOST, 3046\_LARRY\_CROWNE, 3047\_LIFE\_OF\_PI, 3048\_LITTLE\_FOCKERS, 3049\_MORNING\_GLORY, 3050\_MR\_POPPERS\_PENGUINS, 3051\_NANNY\_MCPHEE\_RETURNS, 3052\_NO\_STRINGS\_ATTACHED, 3053\_PARENTAL\_GUIDANCE, 3054\_PERCY\_JACKSON\_LIGHTENING\_THIEF, 3055\_PROMETHEUS, 3056\_PUBLIC\_ENEMIES, 3058\_RUBY\_SPARKS, 3060\_SANCTUM, 3061\_SNOW\_FLOWER, 3062\_SORCERERS\_APPRENTICE, 3063\_SOUL\_SURFER, 3066\_THE\_ADVENTURES\_OF\_TINTIN, 3067\_THE\_ART\_OF\_GETTING\_BY, 3069\_THE\_BOUNTY\_HUNTER, 3070\_THE\_CALL, 3071\_THE\_DESCENDANTS, 3072\_THE\_GIRL\_WITH\_THE\_DRAGON\_TATTOO, 3073\_THE\_GUILT\_TRIP, 3075\_THE\_SITTER, 3076\_THE\_SOCIAL\_NETWORK, 3077\_THE\_VOW, 3078\_THE\_WATCH, 3079\_THINK\_LIKE\_A\_MAN, 3081\_THOR, 3082\_TITANIC1, 3083\_TITANIC2, 3084\_TOOTH\_FAIRY, 3085\_TRUE\_GRIT, 3086\_UGLY\_TRUTH, 3087\_WE\_BOUGHT\_A\_ZOO, 3088\_WHATS\_YOUR\_NUMBER, 3089\_XMEN\_FIRST\_CLASS, 3090\_YOUNG\_ADULT, 3091\_ZOMBIELAND, 3092\_ZOOKEEPER]}}

}

\end{document}